\documentclass[lettersize,journal]{IEEEtran}
\usepackage{algorithmic}
\usepackage{array}
\usepackage[caption=false,font=normalsize,labelfont=sf,textfont=sf]{subfig}
\usepackage{textcomp}
\usepackage{stfloats}
\usepackage{url}
\usepackage{verbatim}
\usepackage{graphicx}

\usepackage{multirow,bbding}
\usepackage{amsmath,amssymb,amsfonts,bm}
\usepackage{threeparttable}

\usepackage{booktabs}

\usepackage{tcolorbox}
\usepackage{mdframed}

\usepackage[numbers,sort&compress]{natbib}

\usepackage{hyperref}
\hypersetup{
  hidelinks
}

\usepackage{fontawesome5}

\usepackage{tikz}
\definecolor{lime}{HTML}{A6CE39}
\DeclareRobustCommand{\orcidicon}{%
    \begin{tikzpicture}
    \draw[lime, fill=lime] (0,0) 
    circle [radius=0.16] 
    node[white] {{\fontfamily{qag}\selectfont \tiny ID}};    \draw[white, fill=white] (-0.0625,0.095) 
    circle [radius=0.007];    \end{tikzpicture}
    \hspace{-2mm}}
\foreach \x in {A, ..., Z}{%
    \expandafter\xdef\csname orcid\x\endcsname{\noexpand\href{https://orcid.org/\csname orcidauthor\x\endcsname}{\noexpand\orcidicon}}
}

\def\BibTeX{{\rm B\kern-.05em{\sc i\kern-.025em b}\kern-.08em
    T\kern-.1667em\lower.7ex\hbox{E}\kern-.125emX}}
\usepackage{balance}
\begin{document}

\title{CRCL: Causal Representation Consistency Learning for Anomaly Detection in Surveillance Videos}

\author{Yang Liu\orcidA{}, 
Hongjin Wang\orcidI{},
Zepu Wang\orcidL{},
Xiaoguang Zhu\orcidJ{},
Jing Liu\orcidB{}, 
Peng Sun\orcidF{}, 
Rui Tang, Jianwei Du,
Victor C.M. Leung\orcidM{}, ~\IEEEmembership{Life Fellow,~IEEE},
Liang Song\orcidG{},~\IEEEmembership{Senior Member,~IEEE}

\thanks{
This work was supported in part by the Ministry of Education - China Mobile Research Fund under Grant MCM2020—J-2, in part by the National Key Research and Development Program of China under Project No. 2024YFE0200700 and Subject No. 2024YFE0200703, in part by the National Natural Science Foundation of China under Grant 62250410368, and in part by the China FAW Joint Development Project. This work was also supported in part by the Specific Research Fund of the Innovation Platform for Academicians of Hainan Province under Grant YSPTZX202314, and in part by the Shanghai Key Research Laboratory of NSAI. The authors sincerely appreciate the editors and anomalous reviewers for their time and efforts. \textit{(Corresponding authors: Jing Liu, Peng Sun and Liang Song.)}
}
}

\markboth{}%
{Yang Liu \MakeLowercase{\textit{\textit{et al.}}}: CRCL: Causal Representation Consistency Learning for Anomaly Detection in Surveillance Videos}

\maketitle

\begin{abstract}

  Video Anomaly Detection (VAD) remains a fundamental yet formidable task in the video understanding community, with promising applications in areas such as information forensics and public safety protection. Due to the rarity and diversity of anomalies, existing methods only use easily collected regular events to model the inherent normality of normal spatial-temporal patterns in an unsupervised manner. Although such methods have made significant progress benefiting from the development of deep learning, they attempt to model the statistical dependency between observable videos and semantic labels, which is a crude description of normality and lacks a systematic exploration of its underlying causal relationships. Previous studies have shown that existing unsupervised VAD models are incapable of label-independent data offsets (e.g., scene changes) in real-world scenarios and may fail to respond to light anomalies due to the overgeneralization of deep neural networks. Inspired by causality learning, we argue that there exist causal factors that can adequately generalize the prototypical patterns of regular events and present significant deviations when anomalous instances occur. In this regard, we propose Causal Representation Consistency Learning (CRCL) to implicitly mine potential scene-robust causal variable in unsupervised video normality learning. Specifically, building on the structural causal models, we propose scene-debiasing learning and causality-inspired normality learning to strip away entangled scene bias in deep representations and learn causal video normality, respectively. Extensive experiments on benchmarks validate the superiority of our method over conventional deep representation learning. Moreover, ablation studies and extension validation show that the CRCL can cope with label-independent biases in multi-scene settings and maintain stable performance with only limited training data available.

\end{abstract}


\section{Introduction}~\label{sec1}

\IEEEPARstart{V}{ideo} Anomaly Detection (VAD) aims to proactively discover anomalous spatial-temporal patterns in surveillance videos to automatically detect abnormal events beyond human expectations, such as violent behaviors, traffic congestion, and industrial accidents, which holds promising application potential in the fields like information forensics, smart cities, and modern manufacturing \cite{liu2024generalized}. However, due to the ambiguity and diversity of the anomalies, as well as the unique high-dimensional characteristics of videos, VAD remains a key task to be further explored in the pattern recognition and video processing community \cite{tip2}. Specifically, in the real world, the notion of anomaly is subjective and relative, i.e., similar behavioral patterns may be classified into different categories under different contextual scenarios or human perceptions, suggesting that it is hardly practical to explicitly define all possible anomalies in advance and finely label the available data \cite{ramachandra2020survey}. In addition, anomalous instances occur much less frequently than normal ones, making it difficult to collect a balanced number of samples for training. Therefore, designing models that can cope with diverse anomalies from the real world with only a few or even no anomalous samples available remains a formidable challenge \cite{tip1}.

Existing Deep Representation Learning (DeepReL) methods typically formulate VAD as an unsupervised out-of-distribution detection task that uses only easily collected regular events to learn the prototypical normal patterns, i.e., normality, with the assumption that models trained with normal videos cannot characterize unseen anomalies \cite{huang2022self}. Inspired by Sparse Representation Learning (SRL) \cite{wu2022self}, we argue that normal samples contain both shared and private semantics, i.e., prototypical features common to various types of regular events and label-independent personalized features possessed by individual negative instances. Due to differences in acquisition devices and environmental conditions, real-world surveillance videos usually include random and diverse data biases, especially scene changes. However, the statistical dependencies established by DeepReL-based VAD models \cite{liu2022learning,huang2022hierarchical,liu2023amp} are usually impotent against such disturbances. Regular events with biases, such as normal behaviors occurring in a new scene, maybe misclassified as anomalies, leading to high false alarm rates. In addition, there is a semantic pattern crossover between regular events and anomalies, so it is difficult for unsupervised VAD to effectively ignore such non-discriminative semantics without having seen the positive samples. An unaffordable consequence is that models trained on only normal samples may effectively reason about abnormal instances in the testing phase due to an overabundance of learned patterns thus leading to missed detections \cite{park2020learning}. They attempt to establish statistical dependencies between spatial-temporal patterns of regular events and the negative labels, rendering the learned representations unable to cope with normal events with scenario bias and diverse anomalous instances in the real world. Existing studies demonstrate that it's difficult for DeepReL-based VAD models to maintain a reasonable balance between effectively representing normal patterns and limited generalization to anomalies \cite{liu2023learning}. In recent years, causality learning \cite{scholkopf2021toward} has been proposed to mine the potential causality of observed data rather than simply modeling the statistical dependencies with the given labels, achieving remarkable success in domain generalization \cite{lv2022causality} and recommendation systems \cite{lin2023survey}. Motivated by this, we introduce causality into unsupervised video normality learning and attempt to utilize causal completeness and independence to learn the causal factors in VAD and construct a scene-robust VAD model. 

\begin{figure}
  \centering
  \includegraphics[width=\linewidth]{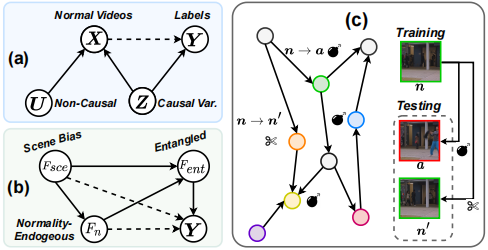}
  \caption{Structural causal models for \textbf{(a)} unsupervised video anomaly detection and \textbf{(b)} scene-debiasing learning from the causality perspective and \textbf{(c)} the schematic diagram of the plausibility analysis of causal representation consistency learning. The CVAD-SCM in \textbf{(a)} highlights the limitations of DeepReL models, which emphasizes establishing the statistical dependencies (dashed arrow) between normal videos $\bm{X}$ and labels $\bm{Y$} but overlook the exploration of causal variable $\bm{Z}$. The Sd-SCM in \textbf{(b)} demonstrates that the deep representation $\bm{F}_{ent}$ is usually an entanglement of the normality-endogenous feature $\bm{F}_n$ and the scene bias $\bm{F}_{sce}$. The sparse mechanism shift shown in \textbf{(c)} posits that label-independent offsets across normal events ($\bm{n}\to \bm{n}^\prime$) exert a limited and localized influence on the learned causality (noted by {\footnotesize \faCut}). Conversely, anomalous events ($\bm{n}\to \bm{a$}) engender an outright breakdown (noted by {\footnotesize \faBomb}) in the inherent consistency.}
  \label{moti}
\end{figure}

To this end, our preliminary research \cite{liu2023learning} has explored causal learning for unsupervised VAD by constructing a Structural Causal Model (denoted as CVAD-SCM), as shown in Fig.~\ref{moti}(a). In CVAD-SCM, $\bm{X}$ and $\bm{Y}$ denote the normal videos and semantic labels in the training set, respectively. Inspired by the Common Causal Principle (CCP), the observable $\bm{X}$ consists of a causal variable $\bm{Z}$ that fully describes the prototypical pattern of regular events, and label-independent non-causal variables $\bm{U}$ caused by variations in acquisition devices and external environments. However, both $\bm{Z}$ and $\bm{U}$ are not directly observable in CausalReL, making it impossible to quantify the impact of scene variations in $\bm{U}$. Therefore, while CVAD-SCM highlights the limitations of DeepReL-based VAD models, it offers limited guidance for constructing scene-robust VAD systems. To address this limitation, this paper proposes a novel SCM for Scene-debiasing (denoted as Sd-SCM) that explicitly models how scene bias $\bm{F}_{sce}$ affects video normality learning through entangled representations $\bm{F}_{ent}$, i.e., $(\bm{F}_{sce}, \bm{F}_n)\to \bm{F}_{ent} \to \bm{Y}$, as shown in Fig.~\ref{moti}(b). Based on this Sd-SCM, we develop Scene-debiasing Learning (SdL) to eliminate scene bias by stripping $\bm{F}_s$ from entangled representations, obtaining discriminative normality-endogenous features $\bm{F}_n$ that follow the ideal path $\bm{F}_n \to \bm{F}_{ent} \to \bm{Y}$. The Sparse Mechanism Shift (SMS) hypothesis \cite{scholkopf2021toward} theoretically justifies our proposed \textbf{C}ausal \textbf{R}epresentation \textbf{C}onsistency \textbf{L}earning (\textbf{CRCL}). As illustrated in Fig.~\ref{moti}(c), diverse normal events exhibit only localized and limited differences in learned causal consistency, despite significant variations in the image domain. During testing, unseen abnormal events $\bm{a}$ with different semantics trigger global changes or collapse of learned causal representation consistency, while new normal events $\bm{n}^\prime$ with unknown label-independent shifts cause only minor changes. The training process integrates Causality-inspired Normality Learning (CiNL) and SdL, optimizing objectives based on Independent Causal Mechanism (ICM) and multi-view pattern consistency. During inference, CRCL identifies anomalies by evaluating whether test samples maintain the learned causal representational consistency.

To conclude, this paper proposes a novel causality-inspired framework named CRCL for scene-robust VAD that makes substantial theoretical advances, technical improvements, and experimental validations. Theoretically, we construct the first Structural Causal Model (SCM) for scene-robust VAD that explicitly describes the causal relationships between scene bias and video normality, and develop a principled approach to analyze and mitigate scene bias through Total Direct Effect. Technically, we treat the spatial-temporal features extracted by traditional DeepReL as entangled representations containing both normality-endogenous features and scene bias, and propose several key innovations: (1) a temporal attention-enhanced encoder $E_m$ to acquire motion-aware representations of raw video clips, which are simultaneously fed into the SdL and CiNL modules to perform scene-debiasing and causal variable mining, respectively; (2) a scene encoder $E_s$ based on 2D convolutional networks to extract background features that are input to SdL for sensing label-independent shifts; and (3) a novel memory addressing mechanism in CiNL to facilitate the separation of shared and private semantics. Experimentally, we demonstrate significant performance improvements in challenging scenarios: CRCL achieves superior results on cross-scene videos, maintains robust performance with limited training samples (even with only 50\% data), and shows consistent advantages across both single-scene and multi-scene settings. These comprehensive advancements in theory, implementation, and validation collectively establish CRCL as a significant step forward in developing a scene-robust VAD model for real-world applications. The main contributions of this paper are summarized as follows:
\begin{itemize}
 \item We develop a causality-based framework for scene-robust VAD, proposing CRCL to learn video normality that implies causal mechanisms and detect real-world video anomalies through representation consistency.
 \item We propose SdL with a structural cause model to strip scene bias from entangled representations, which enables CRCL to resist label-independent shifts.
 \item We introduce temporal attention and memory filtering strategy to CiNL, which enhances CRCL's capability of learning motion dynamics and prototype features.
 \item Extensive experiments validate the superiority of our method, and multi-scene validations confirm its ability to resist scene variations, which can maintain stable performance when even only 50\% training data is available.
\end{itemize}

The structure of the remainder of this paper is outlined as follows. Sec.~\ref{sec2} presents the related work, including the recent advances in VAD and causal learning. Sec.~\ref{sec3} formulates the unsupervised VAD from a causality perspective and presents the principle of scene-debiasing learning. Sec.~\ref{sec6} states the structure, implementation, and optimization process of the proposed CRCL. Sec.~\ref{sec5} provides the results of quantitative experiments, ablation studies, qualitative analysis, and extension discussion. Sec.~\ref{sec6} concludes this paper.

\section{RELATED WORK}~\label{sec2}
\subsection{Video Anomaly Detection}

Early efforts on VAD \cite{chalapathy2018anomaly} follow the unsupervised setting with the open-world assumptions and formulate this task as a one-class classification problem. They attempt to establish boundaries with manually engineered features, which may grapple with the curse of dimensionality. With the development of deep learning, researchers propose to use generative models to learn the spatial-temporal representations and prototypical regular patterns \cite{slavic2022kalman}. For example, ConvAE in \cite{hasan2016learning} reconstructs input sequences with AE and computes anomaly scores based on reconstruction errors. Similarly, Liu \textit{\textit{et al.}} \cite{liu2018future} propose a video prediction framework that quantifies anomaly degrees through prediction errors. Following works include refining structures like the adoption of dual-stream networks \cite{chang2022video,bao2022hierarchical,liu2023amp} to independently learn appearance and motion normality, alongside proxy task aggregation \cite{zhao2017spatio}.

In recent investigations, researchers have explored the intrinsic semantics consistency of normal videos to learn normality, either across different information dimensions \cite{cai2021appearance} or within specific regions \cite{bao2022hierarchical}. AMMC-Net in  \cite{cai2021appearance} utilized memory networks to store prototype patterns of training samples, striving for appearance-motion consistency. Moreover, emerging object-level methods \cite{chen2022comprehensive,bao2022hierarchical} seek to probe the interactions between objects and scene semantics. 
However, these methods risk either failing to detect light anomalies due to overgeneralization or struggling to effectively reason about unseen regular events with data bias owing to restricted representation learning capabilities. 
Compared to the weakly supervised approaches \cite{pu2022locality,wu2024vadclip}, unsupervised VAD struggles to resist label-independent data bias, suffering from significant performance degradation when dealing with multi-scene videos or when only limited regular events are available for training. Although researchers \cite{georgescu2021background} have proposed the use of pre-trained object detection or instance segmentation models to extract foreground objects of interest and individually model their attributes to avoid interference from the background scene, such approaches introduce additional data preprocessing and require predefinition of the targets to be recognized. 
Such methods, including reasoning-based, scene-aware, and scene-dependent approaches, employ explicit reasoning processes to model the associations between foreground objects and background scenes, performing anomaly detection by analyzing semantic deviations in test videos. For example, Sun \textit{et al.} \cite{sun2020scene} developed scene-aware context reasoning, which encodes visual contexts (e.g., object appearance, spatio-temporal relationships, and scene type) as graphs and performs contextual reasoning using recurrent neural networks. Deep evidential reasoning \cite{sun2022evidential} encodes visual cues as evidence and estimates uncertainty based on evidence distributions, utilizing deep Gaussian networks to model foreground-background associations. The hierarchical semantic contrast approach \cite{sun2023hierarchical} leverages pre-trained video parsing models to extract high-level semantic features of foreground objects and background scenes, optimizing feature representations through scene-level and object-level contrastive learning. While these methods demonstrate excellent performance in understanding complex scenes and spatio-temporal interactions, their effectiveness is often constrained by their reliance on object detectors or skeleton extraction models, as well as limitations inherent to data processing tools and computational complexity.
Existing studies \cite{feng2021mist} have shown that weakly supervised methods using video-level labels as human a priori are more reliable and outperform unsupervised methods on multi-scene datasets. However, they can only detect specific anomalies without open-set identification capability. In this regard, we continue to focus on unsupervised VAD and try to revisit this task from a causality perspective to improve the model's ability to adapt to complex scenes. Compared to DeepReL-based models limited by entangled representations and non-causal variables, our proposed CRCL employs CausalReL to remove scene bias and mine causal normality. With the synergy of SdL and CiNL, our CRCL can learn causal factors describing prototype patterns of regular events to strike a balance between representation and generalization. 
\subsection{Causal Learning}

The DeepReL models are mainly centered around learning the statistical dependencies between training samples and given labels, operating under the ideal independent and identically distributed (i.i.d.) paradigm \cite{scholkopf2021toward}. In contrast, causal learning techniques, including causal inference and causal representation learning, attempt to break through the i.i.d restriction in order to enhance the utility of AI algorithms in the real world. Specifically, causal learning treats statistical dependence as an oversimplified abstraction of the physical world, which is deficient in dealing with distributional shifts and intervention scenarios, whereas mining the underlying causality in representation learning will help the model learn the essential connection between data and labels. Causality-driven representation models have shown impressive performance in tasks such as online recommendation systems and domain generalization \cite{lv2022causality}, highlighting the potential of CausalReL to build robust and reusable mechanisms that are essential for achieving scene-robust high-performance VAD.

Real-world videos usually contain unpredictable arbitrary biases that are usually entangled with normality-endogenous features, which may cause DeepReL models to treat them as anomalous cues \cite{liu2024generalized}. In addition, regular events contain individual personalized semantics. DeepReL-based anomaly detectors are usually powerless in adapting to such private features because only regular samples are available during the training phase. Furthermore, the diversity and unboundedness of anomalous events means that their patterns may intersect with those of regular instances. Therefore, it is necessary to obtain representations that accurately describe the essential factors of video normality, which motivates us to introduce CausalReL to decouple entangled representations and learn causal factors. Although there have been attempts to utilize causality in VAD \cite{wu2021learning,lin2022causal}, our CRCL is the first work for unsupervised tasks. Compared to fully unsupervised \cite{lin2022causal} and weakly-supervised tasks \cite{wu2021learning} setting, it is more difficult to construct intervention and representation strategies for unsupervised VAD with only normal samples available. Specifically, Lin \textit{et al.} \cite{lin2022causal} constructed a causal graph and causal inference framework to analyze and eliminate the confounding effects of the pseudo-label generation process in fully unsupervised VAD and utilized counterfactual-based model integration to model long-term event dependencies to improve detection performance. Wu \textit{et al.} focused on temporal cues and feature discrimination in weakly supervised VAD and proposed a causal temporal relationship module to capture local-range temporal dependencies among features to enhance features. Sun \textit{et al.} proposed a causal generative model to distinguish between event-related and event-irrelevant factors in videos, and learned a prototype of event-related factors with a memory enhancement module to eliminate the interference of event-irrelevant factors in anomaly prediction. In contrast, our proposed CRCL takes the first causal look at scene bias in unsupervised video normality learning and attempts to utilize the consistency of normality-endogenous representations to detect anomalies, which performs well on multi-scene datasets and in complex task settings where only a small number of regular events are available for model training.

\section{Preliminary}~\label{sec3}
\subsection{Unsupervised VAD in Causality Perspective}

\begin{figure*}
  \centering
  \includegraphics[width=.98\textwidth]{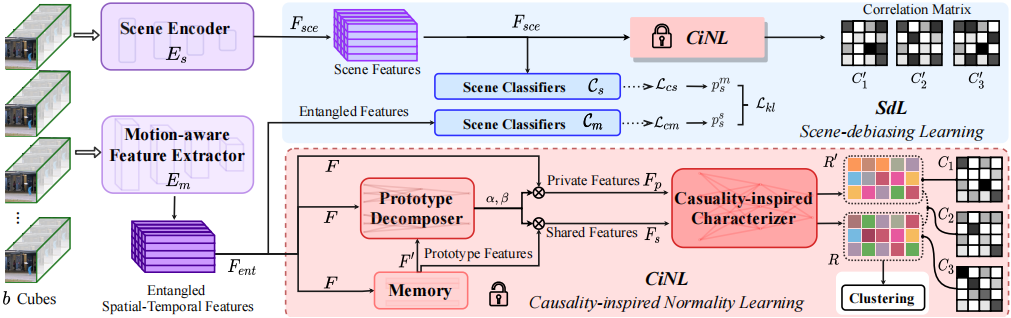}
  \caption{Pipeline overview of the CRCL. SdL utilizes the scene encoder $E_s$ and classifiers $\{\mathcal{C}_s,\mathcal{C}_m \}$ to perceive scene biases in the entangled representation $\bm{F}_{ent}$ from the motion-aware feature extractor $E_m$ and de-biases them from a consistency perspective with the TDE process. In contrast, CiNL consists of a memory network $\mathcal{M}$, prototype decomposer, and CiC to mine the causal variable and utilize a clustering algorithm to obtain task-specific representations.}
  \label{crec}
\end{figure*}

In Sec.~\ref{sec1}, we have stated that spatial-temporal features extracted by DeepReL encapsulate both causal variable that is pivotal in delineating normality and label-independent non-causal variable from unknown bias. Conventional VAD methods \cite{liu2018future,liu2022learning} tend to learn the statistical independence intrinsic to normal videos, which often leads to non-informative representations or over-generalization. To this regard, we introduce a SCM, as illustrated in Fig.~\ref{moti}(a), which guides us to formulate unsupervised video normality learning from the causality perspective and design the CiNL module in Fig.~\ref{crec}. The proposed CRCL aims to explore the causal factors for unsupervised VAD and model robust causal normality beyond the observable training data.

Specifically, we first introduce the CCP \cite{scholkopf2021toward} that unveils the intricate connection between statistical dependence and causality:
\begin{quote}
    \textit{If two observables $\bm{X}$ and $\bm{Y}$ are statistically dependent, then there exists a variable $\bm{Z}$ that causally influences both and explains all the dependence in the sense of making them independent when conditioned on $\bm{Z}$}. 
\end{quote}
CCP usually serves as a fundamental framework in causal inference and causal representation learning, establishing that observed correlations between variables can often be explained by shared causal factors. For the unsupervised VAD task, we introduce the notation $\bm{X}$ for observable regular events and $\bm{Y}$ for video normality (label 0). The unobservable causal variable $\bm{Z}$ occupies a pivotal role, exerting influence both within the original data distribution and the process of normality learning. In other words, $\bm{Z}$ embodies typical targets—counterbalancing unexpected objects in the context of appearance anomalies—or serves as the embodiment of regular object-scene interactions, standing in contrast to violations seen in motion anomalies. With Fig.~\ref{moti}(b), we attempt to implicitly learn causal factors $\{\bm{z}_1, \cdots, \bm{z}_n\}$ from the raw deep representation with internal consistency:
\begin{align}
  & \bm{X}:=f\left(\bm{Z}, \bm{U}, \bm{P}\right), \bm{Z}  \bot \bm{U}  \bot \bm{P}, \\
  & \bm{Y}:=h\left(\bm{Z}, \bm{P}\right)=h\left(g(\bm{X}), \bm{P}^\prime\right), \bm{P}  \bot \bm{P}^\prime,
\end{align}
where $\bm{U}$ symbolizes the non-causal variable that influences solely on $\bm{X}$, representing domain-specific information without relevance to normality learning. In addition, $\bm{P}$ and $\bm{P}^\prime$ denote jointly independent unexplained perturbation noise. The functions $f{(\cdot, \cdot, \cdot)}$, $h(\cdot, \cdot)$, and $g(\cdot)$ constitute unknown structural entities characterized by causal mechanisms. According to CCP and invariant causal mechanism, we know that for any distribution $P(\bm{X}, \bm{Y})\in \mathcal{P}$, the existence of a general conditional distribution $P(\bm{Y}|\bm{Z})$ is a requisite when the causal variable $\bm{Z}$ is provided. Therefore, the representations that inherently convey causality are necessary for robust normality learning that can characterize diverse normal events.

However, direct observation of causal variable for unstructured videos is often unattainable, thereby lacking a guiding precedent to definitively formulate causal representations. In alignment with the established consensus within CausalReL, our CRCL aims to learn a collection of orthogonal causal factors with the ICM \cite{scholkopf2021toward}:
\begin{quote}
    \textit{The causal generative process of a system's variables is composed of autonomous modules that do not inform or influence each other. In the probabilistic case, this means that the conditional distribution of each variable given its causes (i.e., its mechanism) does not inform or influence the other mechanisms.}
\end{quote}
ICM suggests that each causal mechanism operates independently, providing a powerful framework for understanding and modeling complex causal relationships in real-world systems. To be specific, we attempt to unearth $\{\bm{z}_1, \cdots, \bm{z}_n\}$ with their distinct interventions and inherent independence, adhering to the causal factorization premise. Firstly, we deduce that: (1) $\bm{Z}$—completely encapsulating normality—is distinctly separated from $\bm{U}$, thereby establishing that interventions on $\bm{U}$ fail to exert any influence over $\bm{Z}$ and $\bm{Y}$. (2) The collective set of causal factors $\{\bm{z}_1, \cdots, \bm{z}_n\}$ exhibit joint independence, wherein the mechanism $P(\bm{z}_i|PA_i)$ remains unaffected by, and refrains from transmitting information to, $P(\bm{z}_j|PA_j)$ when $j\ne i$. In this context, $PA$ denotes the causal parents. (3) The causal representations tailored for the specific task encapsulate a causal sufficiency, serving as the bedrock for comprehensively explicating all statistical independence shared between $\bm{X}$ and $\bm{Y}$. Finally, the joint distribution of these causal factors are factorized into the conditionals as follows:
\begin{equation}
  P\left(\bm{z}_1, \cdots, \bm{z}_n\right)=\prod_{i=1}^n P\left(\bm{z}_i \mid P A_i\right).
\end{equation}

\subsection{Causality Analysis for Scene-debiasing Learning}

Moreover, CRCL perceives possible scene biases in the entangled representations through causal intervention. Specifically, we first construct the structural causal graph shown in Fig.~\ref{moti}(b) to analyze the potential impact of scene bias on the existing normality learning paradigm, where $\bm{F}_{ent}$ denotes the features of video sequence acquired by the generative models, which are determined by the normality-endogenous features $\bm{F}_n$ and scene bias $\bm{F}_{sce}$. The ideal pipeline of VAD models applicable to real-world applications is $\bm{F}_n\to \bm{F}_{ent}\to \bm{Y}$, i.e., the semantic representations for discriminating anomalies are derived from $\bm{F}_n$ without being affected by $\bm{F}_{sce}$. However, the typical procedure of existing DeepReL-based methods \cite{hasan2016learning,lu2019future,liu2022learning} is $(\bm{F}_n, \bm{F}_{sce}) \to \bm{F}_{ent}\to \bm{Y}$, since that they directly use encoders to extract deep features without taking into account the scene variations, where $\bm{F}_{sce}\to \bm{F}_{ent}\to \bm{Y}$ presents the $\bm{F}_{sce}$ harming the VAD model by affecting the discriminative power of $\bm{F}_{ent}$. They attempt to establish the statistical dependencies (i.e., $\bm{F}_n\to \bm{Y}$ and $\bm{F}_{sce}\to \bm{Y}$), and scene bias affects performance by interfering with the intermediates of representation learning.

Motivated by causal inference, we use do-calculate $do(\bm{X})$ to truncate the effect of $\bm{F}_{sce}$ on normality learning as follows:
\begin{equation}
  \scalebox{0.95}{$
  P(Y \mid \operatorname{do}(\bm{F}_n))=\sum_C P(Y \mid \bm{F}_n, C=\bm{F}_{sce}) P(C=\bm{F}_{sce}),
  $}
\end{equation}
where $P(C=\bm{F}_{sce})$ in the backdoor adjustment requires explicitly scene-specific effects $C$, which is difficult to implement for VAD where only regular events are available in the training phase. In this regard, we simulate the debiasing process at the detection results using TDE, as follows:
\begin{equation}
  \label{eq5}
  \mathrm{TDE} \leftarrow Y_{\bm{F}_n, c}(\bm{F}_{ent})-Y_c(\bm{F}_{ent}),
  \end{equation}
where $Y_{\bm{F}_n, c}$ and $Y_c$ denote the general detection results and the bias-specific results, respectively. By concretizing the effect of $\bm{F}_{sce}$, we can directly remove the negative bias and explore the nature of $\bm{F}_n$ concerning $\bm{F}_{ent}$ through subtraction.

\subsection{Feasibility Analysis}

As stated in Sec.~\ref{sec1}, we regard both shared and private features of regular events as manifestations of normality. Given the expansive range of video events, conventional representation learning faces difficulties in precisely delineating the distribution of these multi-view features, which inherently point towards the same causal variable. It becomes feasible to acquire causal representations through implicit independence, aligning with the SMS \cite{scholkopf2021toward} hypothesis:
\begin{quote}
    \textit{Small distribution changes tend to manifest themselves in a sparse or local way in the causal factorization, that is, they should usually not affect all factors simultaneously.}
\end{quote}
SMS principle suggests that changes in complex systems typically affect only a subset of causal mechanisms, providing a valuable framework for identifying and analyzing distributional changes in causal systems. This propels us towards the acquisition of causal variable that demonstrates heightened responsiveness to anomalies through the intrinsic consistency. To concretely implement this idea, we introduce a prototype decomposer, which partitions the original deep representations into private and shared features for training the causality-inspired characterizer. We introduce clustering techniques and apply cosine similarity constraints to facilitate the exploration of causal consistency, culminating in the derivation of task-specific representations. Furthermore, the TDE process offers a solution for eliminating scene bias, as expressed in Eq.~\ref{eq5}. By mutually optimizing SdL and CiNL, the proposed CRCL can incrementally perceive the entangled biases within the deep representations and remove them through causal interventions.

\section{Methodology}~\label{sec4}
In this section, we first analyze how CiNL utilizes external memory network $\mathcal{M}$ to separate private and shared semantics $\{\bm{F}_p, \bm{F}_s\}$ and utilize a Causality-inspired Characterizer (CiC) to learn the orthogonal causal factors $\{\bm{z}_1, \cdots, \bm{z}_n\}$ with label consistency between $\bm{F}_p$ and $\bm{F}_s$. Then, we introduce two scene classifiers, $\{\mathcal{C}_m, \mathcal{C}_s\}$, to SdL for detecting the scene bias $\bm{F}_{sce}$ in entangled representations $\bm{F}_{ent}$, which eliminate $\bm{F}_{sce}$ through collaborative optimization with CiNL. Finally, we explain how the well-trained CRCL model quantitatively calculates anomaly scores by measuring whether the test samples exhibit causal consistency.

\subsection{Prototype Learning and Decomposition}

\begin{figure}[t]
  \centering
  \includegraphics[width=.98\linewidth]{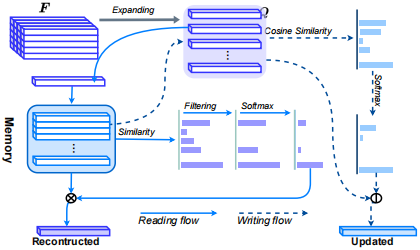}
  \caption{Pipeline of the memory network. We introduce filtering strategy that retains only high-similarity items to reconstruct prototype features. 
  }
  \label{ta}
\end{figure}

The memory can retain prototypes of normal events and curb the tendency of DeepReL to overgeneralize unseen abnormal instances \cite{park2020learning}. As shown in Fig.~\ref{crec}, the memory update process $\mathcal{M}_t \to \mathcal{M}_{t+1}$ epitomizes the assimilation of general spatial-temporal feature $\bm{F}\in \mathbb{R}^{H\times W\times C}$. In particular, the memory pool is depicted as a two-dimensional matrix, symbolized as $\bm{M}\in \mathbb{R}^{C\times N}$, where $N$ signifies the count of memory entries that influence information capacity. Notably, the memory pool remains devoid of learnable parameters, yet exhibits the capability to adaptively update its memory entries to encode prototype features. This is achieved through a write operation employing $\bm{M}$ as a query $\bm{Q}_M$, as expressed below:
\begin{equation}
  \mathcal{M}_{t+1} = l_2\left(\bm{M}+\bm{V}_F \varPsi \left(\frac{\bm{K}_F^T\bm{Q}_M}{\sqrt{C}}\right)\right).
\end{equation}
Here, $\bm{V}_F=\bm{K}_F=e(\bm{F})\in \mathbb{R}^{C\times {N}}$, where $e(\cdot)$ denotes an expansion operation along the spatial dimension, yielding ${N}=H\times W$. The application of $l_2(\cdot)$ maintains a consistent data scale between $\mathcal{M}_t$ and $ \mathcal{M}_{t+1}$, while $\varPsi$ denotes the softmax operation. Conversely, the read operation is used to reconstruct $\bm{F}$ in the form of a prototype $\bm{F}^\prime$, utilizing the expanded $\bm{F}$ as the query $\mathcal{Q}_F$, as follows:
\begin{equation}
  \scalebox{0.95}{$
  \bm{F}^\prime = \bm{V}_M \left| \varPsi \left(\frac{\bm{K}_M^T e(\bm{F})}{\sqrt{C}}\right)\right|, \bm{V}_M=\bm{K}_M=\bm{M}\in \mathbb{R}^{C\times N}.
  $}
\end{equation}
where $\left|\cdot \right|$ denotes a filtering strategy that retains only top-$k$ relevant items for reconstructing $\bm{F}$, as shown in Fig.~\ref{ta}. Compared to using all $N$ memory items to reconstruct $\bm{F}$, the top-$k$ filtering mechanism encourage memory items to focus on typical normal patterns and ignore low-frequency personalized features in $\bm{F}$, making $\bm{F}^\prime$ close to the prototypical features of regular videos. 

\begin{figure}[t]
  \centering
  \includegraphics[width=.98\linewidth]{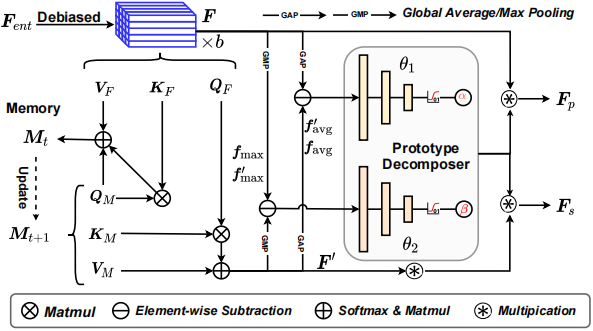}
  \caption{Pipeline of memory-based prototype recording and private-shared features decomposition. The memory network updates memory items ($\mathcal{M}_t \to \mathcal{M}_{t+1}$) to store the prototype features of regular events through attention addressing mechanism, while the prototype decomposer splits private $\bm{F}_p$ and shared features $\bm{F}_s$ with two pooling operations and two MLPs $\{\theta_1, \theta_2\}$.}
  \label{detail}
\end{figure}

The debiased features $\bm{F}$ comprise both shared prototypical semantics and unique personalized semantics, designated as $\{\bm{F}_s, \bm{F}_p\}$. As detailed in Sec.~\ref{sec1}, both $\bm{F}_s$ and $\bm{F}_p$ exhibit statistical affinity with label 1. Follwing SRL \cite{wu2022self}, we utilize a Squeeze-and-Excitation (SE)-like process \cite{hu2018squeeze} to segregate $\bm{F}_s$ and $\bm{F}_p$ from $\bm{F}$ and $\bm{F}^\prime$. This process is illustrated in Fig.~\ref{detail}. Initially, $\bm{F}$ and $\bm{F}^\prime$ undergo average and max pooling, manifesting as $\{\bm{f}_\text{avg},\bm{f}^\prime_\text{avg},\bm{f}_\text{max},\bm{f}^\prime_\text{max}\} \in \mathbb{R}^{C}$. Subsequently, two multi-layer perceptions (MLPs) with learnable parameters $\{\theta_1, \theta_2\}$, convert them into difference scores $\{\alpha, \beta\}$:
\begin{equation}
  \alpha = \text{MLP}(\bm{f}_\text{avg}-\bm{f}^\prime_\text{avg}; \theta_1), \beta = \text{MLP}(\bm{f}_\text{max}-\bm{f}^\prime_\text{max}; \theta_2).
\end{equation}
Finally, we employ $\alpha$ and $\beta$ to separate $\bm{F}_p$ and $\bm{F}_s$ with channel-wise multiplication $\circledast$, as follows:
\begin{equation}
  \bm{F}_p = \frac{\alpha+\beta}{2} \circledast \bm{F}, \quad \bm{F}_s = \left(1-\frac{\alpha+\beta}{2}\right) \circledast \bm{F}^\prime.
\end{equation}

\subsection{Representation Consistency Learning}
Building upon the insights from CCP and ICM, we acknowledge the existence of jointly independent causal factors that have the potential to comprehensively encapsulate the statistical dependencies ranging from low-level video content to high-level normality. Additionally, SMS underscores the notion that distinct features characterizing normal events exert only limited influence over the causal factors and their consistency. Thus, we introduce a CiC that is tailored to learn these unobservable causal variables and to model the intrinsic consistency inherent to normal events. As shown in Fig.~\ref{crec}, we take the spatial-temporal features from a batch of $b$ video clips and decompose them, subsequently feeding these components into the CiC. This process leads to the mapping of shared and private features into causal representations:
\begin{equation}
\scalebox{0.95}{$
\bm{R} = \text{CiC}(\bm{F}_s^1, \bm{F}_s^2, \cdots, \bm{F}_s^b), 
\tilde{\bm{R}} = \text{CiC}(\bm{F}_p^1, \bm{F}_p^2, \cdots, \bm{F}_p^b),
$}
\end{equation}
where $\bm{R}$ and $\tilde{\bm{R}}$ indicate causal representations and share the same dimensionality of $\mathbb{R}^{b \times n}$. In practice, the value of $n$ is substantially smaller than that of $H\times W\times C$. In the pursuit of unsupervised VAD, the objective is to learn normality exclusively from regular events, enabling the causal representations $\bm{r}_i$ and $\tilde{\bm{r}}_i$ to denote the same label. Consequently, the causal variables are anticipated to remain causally invariant when subjected to a decomposition intervention. In essence, the causal representations of shared and private features should remain proximal within the causal factor dimension, as follows:
\begin{equation}
\label{eq8}
\max \frac{1}{n} \sum_{i=1}^n \frac{\langle \bm{f}_i, \tilde{\bm{f}}_i \rangle}{\parallel \bm{f}_i\parallel \parallel \tilde{\bm{f}}_i\parallel},
\end{equation}
where $\bm{f}_i$ and $\tilde{\bm{f}_i}$ stand for the $i$-th column of $\bm{R}$ and $\tilde{\bm{R}}$.

By maximizing the similarity between the same set of $n$ causal factors in shared and private features, we incentivize the CiC to learn causal factors capable of stripping away label-independent non-causal variables. To ensure that the causal factors remain jointly independent, we construct three correlation matrices pertaining to the relationships $\bm{R} \to \tilde{\bm{R}}$, $\bm{R}\to \bm{R}$, and $\tilde{\bm{R}} \to \tilde{\bm{R}}$, denoted as $\bm{C}_1$, $\bm{C}_2$, and $\bm{C}_3$ respectively, as depicted in Fig.~\ref{crec}. Analogous to Eq.~\ref{eq8}, the off-diagonal elements of $\bm{C}_1$ are representative of the cosine similarity between corresponding columns of $\bm{R}$ and $\tilde{\bm{R}}$. Conversely, $\bm{C}_2$ and $\bm{C}_3$ encapsulate the similarity within $\bm{R}$ and $\tilde{\bm{R}}$ as expressed by $\bm{C}_2(i,j) = \frac{\langle\bm{f}_i,\bm{f}_j\rangle}{\parallel \bm{f}_i\parallel \parallel {\bm{f}_j}\parallel}$ and $\bm{C}_3(i,j) = \frac{\langle\tilde{\bm{f}}_i,\tilde{\bm{f}}_j\rangle}{\parallel \tilde{\bm{f}}_i\parallel \parallel \tilde{\bm{f}}_j\parallel}$.
The ultimate optimization objective involves maximizing the diagonal elements of the correlation matrix $\bm{C}_1$ (notably, the diagonal elements of $\bm{C}_2$ and $\bm{C}_3$ remain constant at 1) and minimizing the non-diagonal matrices of $\bm{C}_1$, $\bm{C}_2$, and $\bm{C}_3$. The correlation loss $\mathcal{L}_{c}$ is defined as:
\begin{equation}
\label{eq14}
\mathcal{L}_{c} = \lambda \parallel \bm{C}_1-\bm{I}\parallel_F^2 + \parallel \bm{C}_2-\bm{I}\parallel_F^2 + \parallel \bm{C}_3-\bm{I}\parallel_F^2,
\end{equation}
where $\lambda$ is a hyper-parameter that governs the trade-off of intercorrelation. $\bm{I}$ represents the identity matrix. By optimizing Eq.~\ref{eq14}, we enforce that the causal factors remain jointly independent and resilient to the decomposition intervention. As suggested by SMS, normal events characterized by shifts only induce local differences within the causal representations. Thus, we adopt a clustering approach, similar to \cite{chang2020clustering}, to further refine the causal representation $\bm{R}$, enhancing the model's ability to discriminate among normal events with clustering effects. Additionally, to optimize the memory pool, we follow \cite{park2020learning} to introduce memory separateness and compactness loss.

\subsection{Scene-debiasing Learning}
\begin{figure}[t]
  \centering
  \includegraphics[width=.98\linewidth]{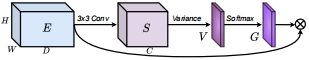}
  \caption{Pipeline of the temporal attention. 
  $E_m$ uses channel variance-based temporal attention to actively capture important motion dynamics.
  }
  \label{tem}
\end{figure}

We propose the SdL to discover the scene bias in the entangled representations, as shown in Fig.~\ref{crec}. In the experimental phase, we utilize multi-scene datasets (e.g., ShanghaiTech \cite{shanghai}) to validate the effectiveness of CRCL and further demonstrate the generalizability of SdL by mixing single-scene datasets \cite{ped2,avenue}. The inputs to CiNL are spatial-temporal representations computed by $E_m$, which aims to capture the motion dynamics with temporal attention \cite{chang2022video}. The computation process is shown in Fig.~\ref{tem}. To be specific, We first compress the input feature map $\bm{E} \in \mathbb{R}^{H\times W \times C}$ into $\bm{S} \in \mathbb{R}^{H\times W \times D}$ via a $3\times 3$ convolution layer, where $D < C$. The temporal variance map $\bm{V} \in \mathbb{R}^{H\times W}$ is then computed through $l_2$ normalization along the channel dimension:
\begin{equation}
    \resizebox{0.97\width}{!}{$
    \bm{V}(i, j)=\frac{1}{D} \sum_{k=1}^{D} \left\|\bm{S}(i, j, k)-\frac{1}{D} \sum_{k=1}^{D} \bm{S}(i, j, k)\right\|_2^2$},
\end{equation}
where $(i,j)$ indexes spatial locations. The attention map $\bm{G} \in \mathbb{R}^{H\times W}$ is obtained through spatial Softmax normalization:
\begin{equation}
    \resizebox{\width}{!}{$
    \bm{G}(i, j)=\left\|\frac{\exp (\bm{V}(i, j))}{\sum_{i=1, j=1}^{H, W} \exp (\bm{V}(i, j))}\right\|_2^2$}.
\end{equation}
The resulting attention-enhanced features emphasize significant motion patterns while suppressing minor temporal variations, effectively capturing motion-relevant information for normality learning. In contrast, SdL focuses on the repetitive background, so we use the simple 2D convolutional network to implement $E_s$ for scene feature extraction.

The entangled feature $\bm{F}_{ent}$ output from $E_m$ and the scene feature $\bm{F}_{sce}$ from $E_s$ are simultaneously fed to SdL to get rid of the scene bias by performing the TDE process show in Eq.~\ref{eq5}. Specifically, we introduce two Multiple Layer Perceptrons (MLPs) as scene classifiers, denoted as $\mathcal{C}_s$ and $\mathcal{C}_m$, to improve $E_s$ to capture scene information and assist CiNL in perceiving the bias. The $\mathcal{C}_s$ takes the $\bm{F}_{sce}$ of $b$ video cubes as input with the cross-entropy loss function $\mathcal{L}_{cs}$:
\begin{equation}
  \mathcal{L}_{cs} = -\frac{1}{b} \sum_{i=1}^b \sum_{j=1}^{N_s} y_{i j} \log \left(p_{i j}\right),
\end{equation}
where $y_{i j}$ denotes the true label for sample $i$ for class $j$ (1 if the sample belongs to the class, otherwise 0). $p_{i j}$ is the predicted probability calculated by $\mathcal{C}_s$ for sample $i$ for scene class $j$. $N_s$ denotes the number of scenes in the training set, which is 13 for ShanghaiTech and 1 for the common single-scene datasets. Similarly, $\mathcal{C}_m$ takes the $\bm{F}_{ent}$ as input and is also optimized with the cross-entropy loss, denoted as $\mathcal{L}_{cm}$. Although we introduce temporal attention in $E_m$ to adaptively capture important temporal information while weakening the background, the acquired spatial-temporal representations still contain label-independent scene information. In response, $\mathcal{C}_m$ fits the scene bias distribution entangled with $\bm{F}_{ent}$ by mutual learning with the following KL loss $\mathcal{L}_{KL}$:
\begin{equation}
  \mathcal{L}_{KL} =  \mathcal{D}_{\mathrm{KL}}\left({\boldsymbol{p}}_s^s \| {\boldsymbol{p}}_s^m\right)+\mathcal{D}_{\mathrm{KL}}\left({\boldsymbol{p}}_s^m \| {\boldsymbol{p}}_s^s\right), 
  \end{equation}
where ${\boldsymbol{p}}_s^m$ and ${\boldsymbol{p}}_s^s$ denotes the scene classification probabilities output by $\mathcal{C}_m$ and $\mathcal{C}_s$, respectively, as shown in Fig.~\ref{crec}. Finally, we model the TDE process in Eq.~\ref{eq5} by comparing the causal correlation matrices $\{\bm{C}_1^\prime, \bm{C}_2^\prime, \bm{C}_3^\prime\}$ and $\{\bm{C}_1, \bm{C}_2, \bm{C}_3\}$, as follows:
Specifically, $\bm{F}_{sce}$ is fed into CiNL, and again after prototype decomposition and causal characterization the causality matrices $\{\bm{C}_1^\prime, \bm{C}_2^\prime, \bm{C}_3^\prime\}$ can be computed, whereas since the scene features are unhelpful for normality learning, therefore, such features do not satisfy the CCP and ICM principle. Therefore, we propose the following triplet loss $\mathcal{L}_{t}$:
\begin{equation}
  \mathcal{L}_{t} =  (d(\{\bm{C}_1, \bm{C}_2, \bm{C}_3\}, \bm{I})-d(\{\bm{C}_1^\prime, \bm{C}_2^\prime, \bm{C}_3^\prime\}, \bm{I})+\alpha, 0)
\end{equation}
where $d(\cdot, \cdot)$ is the distance metric, as shown in Eq.~\ref{eq14}. $\alpha$ is a margin parameter that controls the minimum difference between the positive (i.e., $\bm{C}_{1,2,3}^\prime$ and $\bm{I}$) and negative pairs (i.e., $\bm{C}_{1,2,3}$ and $\bm{I}$). Via the collaborative learning of SdL and CiNL, the CRCL can perceive the scene bias distribution in $\bm{F}_{Fent}$ and gradually erase its negative impact.

\subsection{Anomaly Detection with Causal Consistency}
Given the unique context of the anomaly detection task, where solely negative samples are at our disposal for training, the adeptly trained CRCL finds its efficacy in adeptly decomposing and constructing causal representation consistency for normal events. As we transition into the testing phase, the computation of the anomaly score $s_t$ is founded upon gauging deviations from the learned causal factors, taking into account both their consistency and representations:

\begin{equation}
  \label{eq10}
  s_t = g(\parallel \bm{C}_1-\bm{I} \parallel_F^2 \times D),
\end{equation}
where the function $g(\cdot)$ pertains to the process of max-min normalization, encompassing all frames of a given test video. $D$ signifies the clustering distance spanning the causal representations of the given input video clip and the cluster center. The first part of $s_t$, i.e., $\bm{C}_1-\bm{I}$, discriminates anomalies by virtue of the consistency intrinsic to the causal variable. 

\section{Experiments}~\label{sec5}

\subsection{Experimental Setup}~\label{sec51}

\begin{table}[]
  \centering
  \caption{Details of the unsupervised VAD datasets.}
  \label{dataset}
  \resizebox{.98\linewidth}{!}{
  \begin{tabular}{@{}lccccc@{}}
  \toprule
  \multicolumn{1}{c}{\multirow{2}{*}{\textbf{Dataset}}} & \multicolumn{2}{c}{\textbf{\#Videos}} & \multicolumn{2}{c}{\textbf{\#Frames}} & \multirow{2}{*}{\textbf{\#Scene}} \\ \cmidrule(lr){2-3} \cmidrule(lr){4-5}
  \multicolumn{1}{c}{}       & Train           & Test          & Normal            & Abnormal          &       \\ \midrule
  UCSD Ped2 \cite{ped2}         & 16   & 12 & 2,924             & 1,636             & 1            \\
  CUHK Avenue \cite{avenue}      & 16   & 21 & 26,832            & 3,820             & 1            \\
  ShanghaiTech \cite{shanghai}     & 330  & 107              & 300,308           & 17,090            & 13           \\
  NWPU Campus \cite{cao2023new} & 305  & 242              & 1,400,807           & 65,266            & 43           \\
  \bottomrule
  \end{tabular}
  }
 \end{table}

To validate the efficacy of our proposed CRCL in real-world scenarios, we conduct extensive experiments on three paramount unsupervised VAD datasets. 
Their training sets comprise only normal videos, whereas the anomalous events from the same scene are reserved exclusively for the test sets. The number of videos, frames, and scenes are listed in Table~\ref{dataset}, and the brief introduction is as follows:

\begin{itemize}
  \item \textit{UCSD Ped2} \cite{ped2} is a tiny dataset with 16 training and 12 test videos captured from the university campus. It showcases only one simple scene, with regular instances depicting individuals walking normally along sidewalks, while anomalous events encompass activities like bike riding, skateboarding, and driving.
  \item \textit{CUHK Avenue} \cite{avenue} is also a single-scene VAD dataset, in which the training and test sets encompass 16 and 21 videos, respectively, featuring 47 instances of anomalous events. The dataset designers have meticulously simulated various types of anomalies, including those rooted solely in appearance (e.g., an individual on a lawn), motion (e.g., loitering), and a combination of both (e.g., papers being scattered). To present the applicability of CRCL for multi-scenario VAD, we conduct extended experiments on the mixed dataset of CUHK Avenue and UCSD Ped2 \cite{ped2}.
  
  \item \textit{ShanghaiTech} \cite{shanghai}, as the most practical benchmark within our research scope, encompasses 130 anomalies distributed across 13 distinct scenes. Existing methods either train separate models to handle different scenes, which is of high computation cost and impractical, or ignore the scene differences and treat ShanghaiTech as a single scene dataset. In this work, we not only follow the existing settings to conduct experiments without scene labels to validate the CiNL module but also introduce SdL to test CRCL on multi-scene videos.
  
  \item \textit{NWPU Campus} \cite{cao2023new} is the largest unsupervised VAD dataset, containing 28 types of anomalous events captured from 43 scenarios, such as Climbing fence, Jaywalking, and Littering, as well as scenarios-associated anomalies like Cycling on footpath and Wrong turn. The dataset consists of 305 training videos and 242 test videos, totaling 76.6 GB, with a duration of 16 hours. The largest data size, multi-scene settings, and scene-related anomaly definitions make it the most challenging unsupervised VAD benchmark available.
\end{itemize}

\subsubsection{Evaluation Metrics}
In the testing phase, we assess the input videos by leveraging the learned causality-inspired normality, which yields a continuous anomaly score confined within the interval $[0, 1]$. Notably, a higher score corresponds to a higher likelihood of the tested frame being categorized as anomalous. We proceed to compute true positive rate and false positive rate values at varying thresholds, consequently plotting the receiver operating characteristic curve. The Area Under the Curve (AUC) is used as the primary metric. Furthermore, we incorporate the Equal Error Rate (EER) as an auxiliary metric, underscoring the CRCL's robustness. Moreover, we also report the average inference speed on the CUHK Avenue \cite{ped2} dataset and compare it with state-of-the-art (SOTA) methods on the same implementation platform. 

\subsubsection{Implementation Details}
Our method is implemented using the PyTorch framework and executed on 4 Nvidia 3090 GPUs. The overall training process follows a progressive end-to-end strategy that consists of two main phases. In the initial phase, we first optimize the feature extractor and representation learning networks for 100 epochs without incorporating clustering constraints. This phase aims to establish robust basic feature representation capabilities. Subsequently, in the joint optimization phase, we introduce clustering constraints and employ an alternating optimization technique \cite{chang2020clustering} to compute and update clustering centers through K-means, while simultaneously training the SdL and CiNL modules to optimize the overall objective function. Throughout both phases, the training process is realized by the Adam optimizer with an initial learning rate of $8\times 10^{-5}$, and the batch size $b$ is set to 8. The loss function weights remain fixed during training, and all modules share the same learning rate. 

We resize video frames to dimensions of $224\times 224$ pixels. The $E_m$ is implemented with temporal attention \cite{chang2022video} and 3D convolutional networks while the scene encoder is a 4-layer 2D network. The prototype decomposer consists of two three-layer fully connected neural networks, incorporating a sigmoid activation function within the output layer. For the backbone of CiC, we explore two options: ResNet-18 and ResNet-50 \cite{he2016deep}, denoted as CRCL$^{18}$ and CRCL$^{50}$, respectively. The selection of hyperparameters is guided by empirical insights. Specifically, we set the trade-off hyperparameter $\lambda$ in Eq.~\ref{eq14} to 10, 18, and 20 for the UCSD Ped2 \cite{ped2}, CUHK Avenue \cite{avenue}, and ShanghaiTech \cite{shanghai} datasets, respectively, and the $k$ in the read operation of memory network are 8, 8, and 24.

\begin{table}[]
  \centering
  \caption{Results of the frame-level AUCs on single-scene datasets.}
  \label{t1}
   \resizebox{\linewidth}{!}{
  \begin{threeparttable} 
  \begin{tabular}{@{}clcc@{}}
  \toprule
  \multirow{2}{*}{\textbf{Type}}   & \multicolumn{1}{c}{\multirow{2}{*}{\textbf{Method}}} & \multicolumn{2}{c}{\textbf{Frame-level AUC (\%)}} \\ \cmidrule(l){3-4} 
 & \multicolumn{1}{c}{}  & UCSD Ped2& CUHK Avenue \\ \midrule
  \multirow{6}{*}{{Traditional}}& MPPCA \cite{kim2009observe} & 69.3   & -   \\
 & MPPC+SFA \cite{kim2009observe} & 61.3   & -   \\
 & MDT \cite{ped2}  & 82.9   & -   \\
 & AMDN \cite{xu2017detecting} & 90.8   & -   \\
 & Unmasking \cite{tudor2017unmasking} & 82.2   & 80.6   \\
 & MT-FRCN \cite{hinami2017joint}  & 92.2   & -   \\
 \midrule
  \multirow{20}{*}{{DeepRel-based}} 
 & ConvAE \cite{hasan2016learning}   & 90.0 & 70.2 \\
 & ConvLSTM-AE \cite{lu2019future} & 88.1   & 77.0   \\
 & AMC \cite{nguyen2019anomaly} & 96.2   & 86.9 \\
 & FFP \cite{liu2018future}  & 95.4   & 85.1 \\
 & MemAE \cite{gong2019memorizing} & 94.1   & 83.3 \\
 & SACR \cite{sun2020scene} & -   & 89.6 \\
 & AnoPCN \cite{ye2019anopcn}   & 96.8   & 86.2 \\
 & Mem-Guided \cite{park2020learning} & 97.0 & 88.5 \\
 & AMMC-Net \cite{cai2021appearance}  & 96.6   & 86.6 \\
 & Clustering \cite{chang2020clustering} & 96.5   & 86.0  \\
  & Multi-space \cite{zhang2020normality} & 95.4   & 86.8 \\
 & TAC-Net \cite{huang2021abnormal} & 98.1   & 88.8 \\
 & STD \cite{chang2022video} & 96.7   & 87.1 \\
 & STC-Net \cite{zhao2022exploiting}  & 96.7   & 87.8 \\
  & DEDDnet \cite{zhong2022bidirectional} & 98.1   & 89.0 \\
  & DLAN-AC \cite{yang2022dynamic} & 97.6   & 89.9  \\
 & STM-AE \cite{liu2022learning}  & 98.1   & 89.8  \\
 & ER \cite{sun2022evidential}  & 97.1   & 92.7  \\
 & Bi-Prediction \cite{chen2022comprehensive}   & 97.4   & 86.7 \\
 & HSNBM \cite{bao2022hierarchical} & 95.2   & 91.6 \\
 & MAAM-Net \cite{wang2023memory} & 97.7   & 90.9  \\
 & USTN-DSC \cite{yang2023video} & 98.1   & 89.9  \\
 & GbCL-P \cite{qiu2024video} & 92.2   & 86.2  \\
 & GbCL-R \cite{qiu2024video} & 90.8   & 83.1  \\
 & Trinity \cite{yang2024context} & 97.9   & 88.5 \\
 & A2D-GAN \cite{singh2024attention} & 97.4   & 91.0 \\
 & C3DSU \cite{lappas2024dynamic} & 98.5   & 90.4 \\
 & SSPTL \cite{yang2025video} & \textbf{99.1}   & 91.9 \\
\midrule
\multirow{4}{*}{{CausalReL-Based}} 
 
 & Basic$^{18}$    & 97.6   & 90.5 \\
 & Basic$^{50}$  & 98.7  & 92.5 \\ 
 & CRCL$^{18}$  & 97.8   & 91.2 \\
 & CRCL$^{50}$   & \textbf{99.1}   & \textbf{92.9} \\

 \bottomrule
  \end{tabular}
\begin{tablenotes} 
    \item \small \noindent Bold numbers indicate the best performance on each dataset.
\end{tablenotes} 
\end{threeparttable} 
   }
\end{table}

\subsection{Quantitative Comparison}~\label{sec42}

To validate the effectiveness of the proposed CRCL on real-world surveillance videos, we conduct extensive quantitative experiments on mainstream unsupervised VAD benchmarks, including single-scene videos \cite{ped2,avenue} without taking scene variations into account and multi-scene dataset \cite{shanghai} validation with/without scene labels. In addition, we compare the average inference speed of CRCL with existing methods on the same implementation platform and dataset to demonstrate its deployment potential in real-world applications.

\subsubsection{Performance on Single-Scene Datasets} Table~\ref{t1} shows the frame-level AUCs of CRCL versus traditional manual feature-based and DeepReL-based methods on two single scene VAD datasets: UCSD Ped2 \cite{ped2} and CUHK Avenue \cite{avenue}. Since all videos in such datasets are captured from the same scenario and thus scene-debiasing learning is not applicable. In other words, we only use the CiNL module to learn the normality of regular events. Our CRCL$^{50}$ achieves 99.1\% and 92.9\% AUCs on UCSD Ped2 \cite{ped2} and CUHK Avenue \cite{avenue}, respectively, significantly outperforming traditional methods and existing deep learning approaches. Compared to the prior Basic methods \cite{liu2023learning}, we propose the temporal attention to efficiently capture motion dynamics and introduce a filtering strategy in the memory addressing mechanism to improve the representativeness of the recorded prototype features. In the case of using the same residual network to implement the CiC, CRCL$^{18}$ achieves a performance gain of 0.2\% and 0.7\% on the two datasets, respectively, compared to Basic$^{18}$, which suggests that such improvements for deep representation learning are effective for videos of simple scenes. In addition, the models implemented by ResNet-50 outperform those by ResNet18 for both the Basic \cite{liu2023learning} and the improved CRCL, suggesting that complex networks are more effective in mining the causal factors of unsupervised video normality.
\begin{table}[t]
  \centering
  \caption{Results of the AUCs on the ShanghaiTech \cite{shanghai} dataset.}
  \label{t2}
  \resizebox{.98\linewidth}{!}{
  \begin{threeparttable} 
  \begin{tabular}{@{}lclclccc@{}}
  \toprule
  \textbf{Method} & \textbf{AUC (\%)} & \textbf{Method} & \textbf{AUC (\%)} \\ 
  \midrule

  stackRNN \cite{shanghai} & 68.0 & FFP \cite{liu2018future} & 72.8 \\
  MemAE \cite{gong2019memorizing} & 71.2 & AnoPCN \cite{ye2019anopcn} & 73.6 \\
  Mem-Guided \cite{park2020learning} & 70.5 & AMMC-Net \cite{cai2021appearance} & 73.7 \\
  Clustering \cite{chang2020clustering} & 73.3 & Multi-space \cite{zhang2020normality} & 73.6 \\
  TAC-Net \cite{huang2021abnormal} & 76.5 & STD \cite{chang2022video} & 73.7 \\
  STC-Net \cite{zhao2022exploiting} & 73.1 & DEDDnet \cite{zhong2022bidirectional} & 74.5 \\
  SACR \cite{sun2020scene} & 74.7 & ER \cite{sun2022evidential} & 79.3 \\
  DLAN-AC \cite{yang2022dynamic} & 74.7 & STM-AE \cite{liu2022learning} & 73.8 \\
  Bi-Prediction \cite{chen2022comprehensive} & 73.6 & HSNBM \cite{bao2022hierarchical} & 76.5 \\
  MAAM-Net \cite{wang2023memory} & 71.3 & USTN-DSC \cite{yang2023video} & 73.8 \\
  Trinity \cite{yang2024context} & 74.1 & A2D-GAN \cite{singh2024attention} & 74.2 \\
  SSPTL \cite{yang2025video} & 81.1 & C3DSU \cite{lappas2024dynamic} & 74.3 \\
  \midrule
  Basic$^{18}$ & 77.6 & Basic$^{50}$ & 78.3 \\ 
  CRCL$^{18}$ & 75.4 & CRCL$^{50}$ & 76.1 \\ 
  \textit{CRCL$^{18\dagger}$} & 79.3 & \textit{CRCL$^{50\dagger}$} & \textbf{81.7} \\ 
  \bottomrule
  \end{tabular}
\begin{tablenotes} 
    \item \small $^{\dagger}$Italics indicate that models use scene labels for training.
\end{tablenotes} 
\end{threeparttable} 
   }
\end{table}

\subsubsection{Validation on multi-scene dataset} In addition, we carry out experiments on the multi-scene ShanghaiTech dataset \cite{shanghai} with and without scene labels. The quantitative comparisons of CRCL variants with SOTA DeepReL methods are shown in Table~\ref{t2}. Firstly, we follow the task setting of the existing unsupervised methods that consider ShanghaiTech as a single scene dataset and train the CiNL module directly without using scene labels. Our methods (CRCL$^{18}$ and CRCL$^{50}$ w/o SdL) achieves frame-level AUCs of 75.4\% and 76.1\%, respectively, both lower than the prior methods \cite{liu2023learning} with the same CiC structure, i.e. Basic$^{18}$ and Basic$^{50}$. Compared to the AUC gains of CRCL on single-scene datasets, as shown in Table~\ref{t1}, such decreases indicate that ignoring scene bias and focusing only on improving deep representation learning is not feasible for complex multi-scene videos. Although the CiNL in CRCL introduces components that have worked well in deep representation learning, such as temporal attention \cite{chang2022video} and filtering strategy \cite{liu2023amp}, the presence of scene bias in entangled features prevents the model from discriminative representations, which leads to computational complexity increase and performance degradation. Fortunately, after applying SdL together with the CiNL to remove the scene bias, our CRCL finally achieves the best performance of 81.7\% frame-level AUC at ShanghaiTech \cite{shanghai}. Such a setup requires only ready-made scene labels that do not need to be individually labeled manually, and achieves substantial performance gains without the expensive cost of repeatedly training multiple models. Compared to the SOTA deep learning approach \cite{bao2022hierarchical}, the proposed CRCL achieves an AUC gain of 5.2\%, almost all contributed by SdL. In summary, the CausalReL-based approaches generally outperform the DeepReL-based methods in both training settings, while the SdL further utilizes causal inference to erase data bias from the entangled representation, which in conjunction with the improved CiNL to enhance the capability of CRCL in multi-scene video anomaly detection.

To further evaluate CRCL's potential in real-world surveillance scenarios, we conduct extensive experiments on the challenging NWPU Campus dataset \cite{cao2023new}. The comparative results are presented in Table~\ref{nwpu}, including state-of-the-art methods FBAE \cite{cao2023new} and SSAE \cite{cao2024scene} proposed by the dataset creators. As discussed in Section~\ref{sec51}, NWPU Campus represents a significant advancement in VAD benchmarking, featuring unprecedented scale, scene diversity, and scene-dependent anomalies. While deep representation learning-based methods \cite{park2020learning,zaheer2020old,cai2021appearance} have demonstrated strong performance on standard benchmarks such as UCSD Ped2 \cite{ped2}, CUHK Avenue \cite{avenue}, and ShanghaiTech \cite{shanghai}, their frame-level AUC scores typically fall below 65\% on this more challenging dataset. Our proposed CRCL achieves a frame-level AUC of 69.7\% when treating the multi-scene NWPU Campus videos as a unified training set. The effectiveness of scene-debiasing learning is particularly evident: compared to variants without scene labels (CRCL$^{18}$ and CRCL$^{50}$), the module yields substantial performance improvements of 3.1\% and 6.2\%, respectively. These gains empirically validate the significant impact of scene bias on model performance in scene-dependent VAD tasks. While SSAE \cite{cao2024scene} achieves higher detection accuracy, CRCL$^{50}$ demonstrates superior computational efficiency with an average inference speed of 27 FPS under optimal settings. 

\begin{table}[t]
  \centering
  \caption{Results of the AUCs on the NWPU Campus \cite{cao2023new} dataset.}
  \label{nwpu}
  \resizebox{.9\linewidth}{!}{
  \begin{tabular}{@{}lclc@{}}
  \toprule
  \textbf{Method} & \textbf{AUC (\%)} & \textbf{Method} & \textbf{AUC (\%)} \\ \midrule
  MemAE \cite{gong2019memorizing}          & 61.9              & Mem-Guided \cite{park2020learning}     & 62.5              \\
  OG-Net \cite{zaheer2020old}         & 62.5              & AMMC-Net \cite{cai2021appearance}       & 64.5              \\
  MPN \cite{lv2021learning}            & 64.4              & HF$^2$-VAD \cite{liu2021hybrid}        & 63.7              \\
  FBAE \cite{cao2023new}           & 68.2              & SSAE \cite{cao2024scene}            & \textbf{75.6}              \\ 
  \midrule
  Basic$^{18}$ & 66.9 & Basic$^{50}$ & 67.4 \\ 
  CRCL$^{18}$ & 66.8 & CRCL$^{50}$ & 67.6 \\ 
  \textit{CRCL$^{18\dagger}$} & 70.1 & \textit{CRCL$^{50\dagger}$} & 73.8 \\ 
  \bottomrule
  \end{tabular}
  }
  \end{table}

  \subsubsection{Equal Error Rate and Inference Speed} 
\begin{figure}[t]
  \centering
  \includegraphics[width=\linewidth]{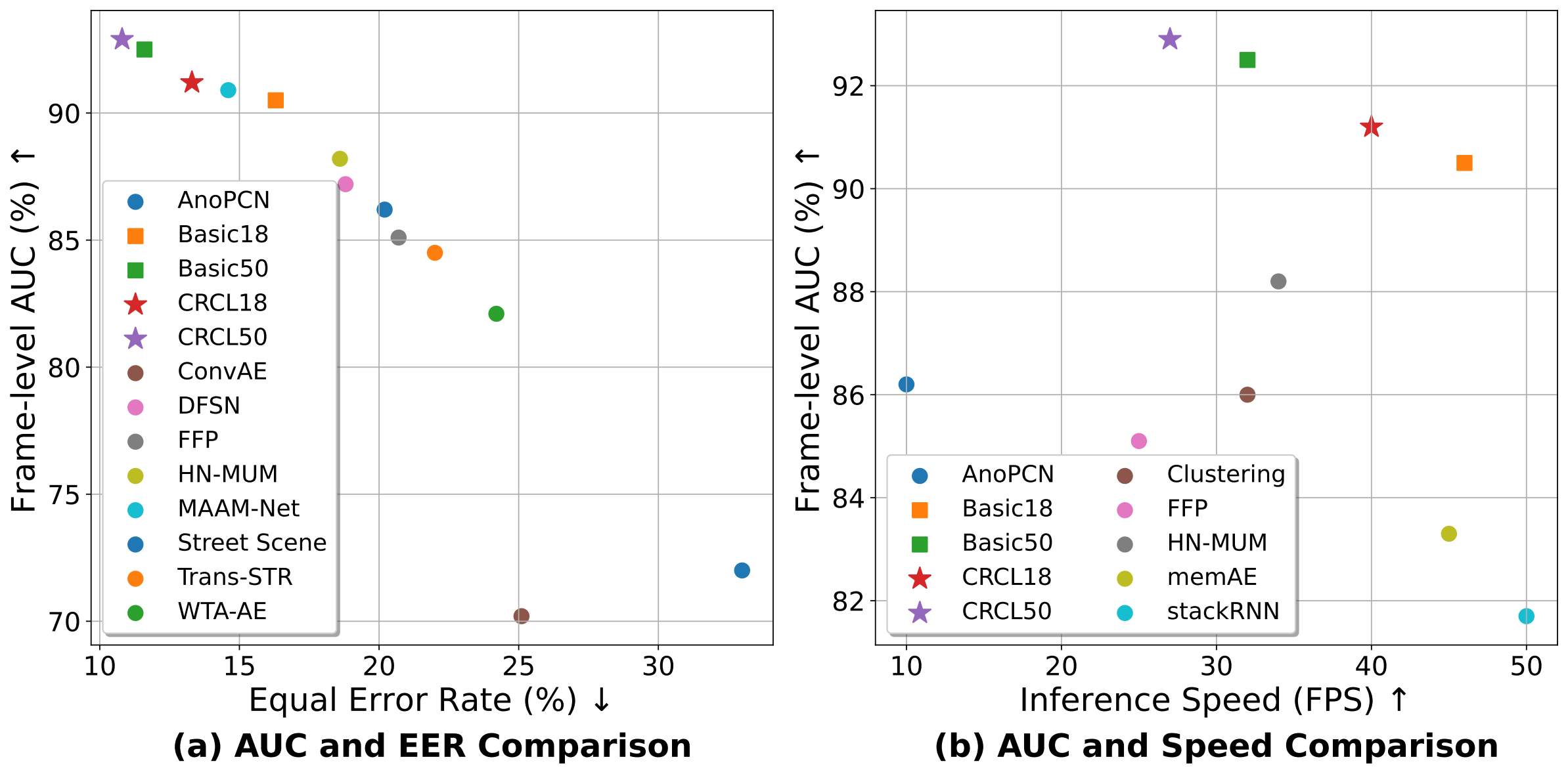}
  \caption{Quantitative EER and inference speed comparison on CUHK Avenue dataset \cite{avenue}. Panel (a) depicts the frame-level AUC and EER, and panel (b) shows the average inference speed. The proposed CRCL is denoted by pentagrams, while the others are represented by squares (Basic methods) and circles. $\uparrow$ indicates that larger values correspond to superior performance, while $\downarrow$ indicates the opposite. The optimal viewing experience in color.}
  \label{scatter}
\end{figure}

Fig.~\ref{scatter} demonstrates the EERs and average inference speed of the proposed CRCL compared to the Basic \cite{liu2023learning} and mainstream deep learning methods on the CUHK Avenue \cite{avenue} dataset. In addition to the methods mentioned in Tables~\ref{t1} and~\ref{t2}, other available comparison models include WTA-AE \cite{tran2017anomaly}, DFSN \cite{ramachandra2020learning}, Street Scene \cite{ramachandra2020street}, Trans-STR \cite{sun2023transformer}, and HN-MUM \cite{li2023hn}. In terms of detection performance, CRCL$^{50}$ achieves the best performance of 10.8\% EER and 92.9\% AUC. For the DeepReL methods with reported values, the lowest EER is 14.6\% achieved by MAAM-Net \cite{wang2023memory}, and our method further significantly reduces it by 26\%, demonstrating the superiority of causal consistency learning in processing large-scale videos. In addition, results on UCSD Ped2 \cite{ped2} and ShanghaiTech \cite{shanghai} also prove the above advantages, with EERs of 3.2\% and 14.1\%, respectively, which are lower than those of the concurrent methods. Although our CRCL lags behind several methods \cite{shanghai,park2020learning} in speed, as shown in Fig.~\ref{scatter}(b), the dominant advantage in detection performance can still indicate its competitiveness. Quantitatively, the average inference speeds of CRCL$^{18}$ and CRCL$^{50}$ are 40 FPS and 27 FPS, which satisfy the real-time detection demand. The processing time from video read-in to anomaly score output is only 0.025s and 0.037s, respectively, which is faster than the human eye. It is worth mentioning that SdL works only in the training phase and is used to assist CiNL in obtaining normality-endogenous features, while in the testing phase, the components for SdL are not involved in the computation of anomaly scores. Therefore, SdL does not incur additional inference costs for model deployment and this speed measurement is applicable to multi-scene videos.

\begin{table*}[t]
  \centering
  \caption{Results of ablation study on components and constraints.}
  \label{t3}
\resizebox{\textwidth}{!}{
  \begin{tabular}{@{}ccccccccccc@{}}
  \toprule
  \multirow{2}{*}{\textbf{ID}} & \multicolumn{5}{c}{\textbf{Component}}& \multicolumn{3}{c}{\textbf{Constraint}} & \multicolumn{2}{c}{\textbf{Frame-level AUC (\%)}} \\ 
  \cmidrule(l){2-6} \cmidrule(l){7-9} \cmidrule(l){10-11}
 & Clustering & Filtering & Temporal Attention & Average Pooling & Max Pooling & $\parallel \bm{C}_1-\bm{I}\parallel_F^2$ & $\parallel \bm{C}_2-\bm{I}\parallel_F^2$ & $\parallel \bm{C}_3-\bm{I}\parallel_F^2$ & UCSD Ped2& CUHK Avenue\\ \midrule
  1    & \XSolidBrush  & \CheckmarkBold & \CheckmarkBold  & \CheckmarkBold & \CheckmarkBold   & \CheckmarkBold   & \CheckmarkBold   & \CheckmarkBold   & 92.3   & 84.4     \\
  2    & \CheckmarkBold  & \XSolidBrush & \CheckmarkBold  & \CheckmarkBold & \CheckmarkBold   & \CheckmarkBold   & \CheckmarkBold   & \CheckmarkBold   & 98.1   & 90.9     \\
  3    & \CheckmarkBold  & \CheckmarkBold & \XSolidBrush  & \CheckmarkBold & \CheckmarkBold   & \CheckmarkBold   & \CheckmarkBold   & \CheckmarkBold   & 98.5   & 91.8     \\
  4    & \CheckmarkBold  & \CheckmarkBold & \CheckmarkBold  & \XSolidBrush & \CheckmarkBold   & \CheckmarkBold   & \CheckmarkBold   & \CheckmarkBold   & 98.3   & 91.5     \\
  5    & \CheckmarkBold  & \CheckmarkBold & \CheckmarkBold  & \CheckmarkBold & \XSolidBrush   & \CheckmarkBold   & \CheckmarkBold   & \CheckmarkBold   & 98.5   & 91.9     \\
  \midrule
  6    & \CheckmarkBold  & \CheckmarkBold & \CheckmarkBold  & \CheckmarkBold & \CheckmarkBold   & \XSolidBrush   & \CheckmarkBold   & \CheckmarkBold   & 90.1   & 83.6     \\
  7    & \CheckmarkBold  & \CheckmarkBold & \CheckmarkBold  & \CheckmarkBold & \CheckmarkBold   & \CheckmarkBold   & \XSolidBrush   & \CheckmarkBold   & 98.0     & 90.6     \\
  8    & \CheckmarkBold  & \CheckmarkBold & \CheckmarkBold  & \CheckmarkBold & \CheckmarkBold   & \CheckmarkBold   & \CheckmarkBold   & \XSolidBrush   & 98.3   & 91.4     \\
  9    & \CheckmarkBold  & \CheckmarkBold & \CheckmarkBold  & \CheckmarkBold & \CheckmarkBold   & \CheckmarkBold   & \XSolidBrush   & \XSolidBrush   & 97.5   & 90.2     \\
  \midrule
  10   & \CheckmarkBold  & \CheckmarkBold & \CheckmarkBold  & \CheckmarkBold & \CheckmarkBold   & \CheckmarkBold   & \CheckmarkBold   & \CheckmarkBold   & 99.1   & 92.9     \\ \bottomrule
  \end{tabular}
}
\end{table*}

\subsection{Ablation Study}

The comparison experiments on single-scene datasets have demonstrated the superiority of CRCL over existing deep learning methods and our prior work \cite{liu2023learning}, and the validation on multi-scene video (see Table~\ref{t2}) presents the effectiveness of scene-debiasing learning. To further explore the impact of individual components and optimization constraints on causal video normality learning, we conduct extensive ablation studies on the UCSD Ped2 \cite{ped2} and CUHK Avenue \cite{avenue} datasets. The  results are shown in Table~\ref{t3}, and the detailed analysis is as follows:

\subsubsection{Analysis of the Effect of Components} We quantitatively present the effectiveness of the key components by removing them step by step and calculating the frame-level AUCs of the variant models with the same experimental setting. Specifically, the Model 1 in Table~\ref{t3}, which removes the clustering module and uses only causality-inspired normality learning, exhibits significant performance degradation on both single-scene datasets. As mentioned in Sec.~\ref{sec1}, unsupervised VAD follows the one-class classification setting, and thus the operation in existing CausalReL models, e.g., constructing classifiers with goals of downstream task, is not applicable to our task. In this work, we utilize the clustering algorithm to encourage causal representations of regular events to aggregate with each other and empower CRCL to learn task-specific representations. During the testing phase, the well-trained CRCL assesses anomalies by using the distance of the input samples from the nearest clustering center. Thus, the clustering module and its co-optimization process with CiNL are crucial. Compared to the Basic methods \cite{liu2023learning}, CRCL introduces a filtering strategy and temporal attention in CiNL's memory network and the motion-aware feature extractor $E_m$, respectively, in addition to the introduction of SdL for overcoming the scene bias to cope with multi-scene videos. The AUC gap of Model 2 over Model 10 suggests that the filtering strategy enhances the ability to learn prototypical features, which can improve model performance gains on complex datasets. In contrast, temporal attention aims to improve the discriminative power of entanglement representations by enhancing $E_m$'s ability to capture motion dynamics, but the presence of data bias makes its contribution limited, as illustrated by Model 3. Moreover, Models 4 and 5 compare the impact of average and maximum pooling in the prototype decomposer. The apparent AUC gap between the two highlights the efficacy of average pooling in aggregating global information and separating shared and private features more efficiently. Furthermore, both pooling strategies contribute to consistency learning and show cumulative gains compared to the full CRCL in Model 10.

\subsubsection{Effectiveness Analysis of Optimization Strategies:} According to CCP and ICM, we optimize the CiC module to mine causal factors in video normality learning by imposing constraints on the correlation matrix, as expressed in Eq.~\ref{eq14}. The results in Models 6-9 in Table~\ref{t3} reveal the quantitative impact of each optimization term on the overall performance. Among them, Model 6 suffers the most significant performance degradation, even lower than earlier DeepReL-based baseline methods. In contrast, Model 10 achieves frame-level AUC improvements of 9.0\% and 9.3\% over Model 6 on UCSD Ped2 \cite{ped2} and CUHK Avenue \cite{avenue}, respectively, just by applying the constraints on the correlation matrix of $\bm{R} \to \hat{\bm{R}}$. This demonstrates the decisive role of $\parallel \bm{C}_1-\bm{I}\parallel_F^2$ in causal representation learning, and validates the effectiveness of the independent causal mechanism. In addition, the impact of the other two constraint items of the inner-correlation matrix, i.e., $\parallel \bm{C}_2-\bm{I}\parallel_F^2$ in Model 7 and $\parallel \bm{C}_3-\bm{I}\parallel_F^2$ in Model 8, is limited. Model 9 directly removes the above two constraints to further reduce the computation costs, which suffers an acceptable performance decrease and the performance is still satisfactory for real-world detection.

\subsection{Qualitative Analysis}

\begin{figure}[t]
  \centering
  \includegraphics[width=\linewidth]{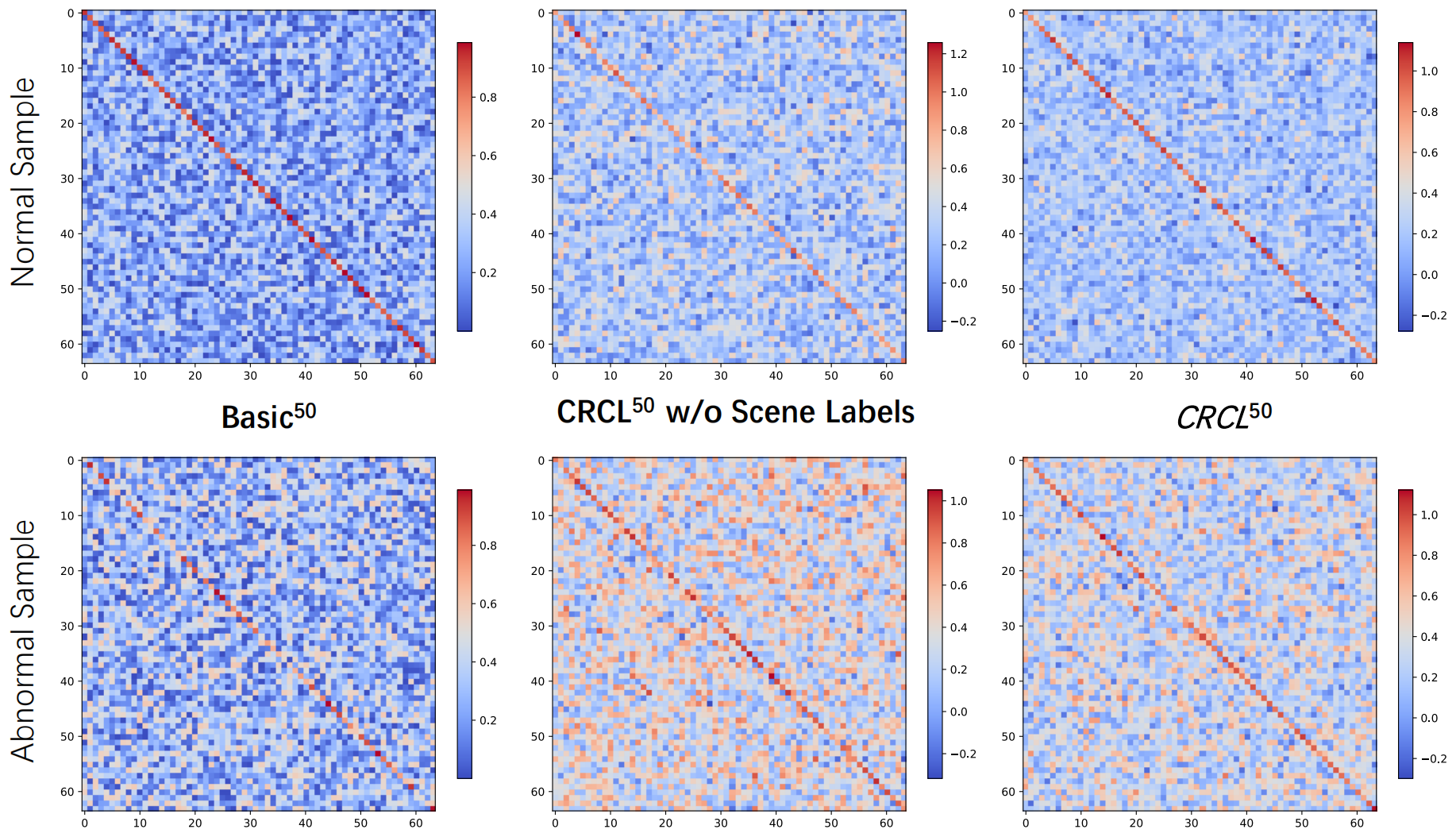}
  \caption{Correlation matrix visualization. We show the $\bm{R}\to \tilde{\bm{R}}$ correlation matrices of Basic$^{50}$ \cite{liu2023learning}, CRCL$^{50}$ without SdL, and full CRCL$^{50}$ on the normal and abnormal test samples from ShanghaiTech dataset \cite{shanghai}.}
  \label{correlation}
\end{figure}

\subsubsection{Correlation Matrix Visualization} Our CRCL aims to explore potential causal variable of unsupervised VAD tasks with causal representation learning and to detect anomalies using the learned causal representation consistency. It is feasible in terms of learning causal factors that are robust to label-independent data shifts for regular events and exhibit drastic changes in the face of anomalous instances. To visualize the responses of causal representations when encountering normal and abnormal samples and to explain how CRCL performs anomaly detection with consistency, we randomly select a pair of samples from the test set of ShanghhaiTech \cite{shanghai} and visualize their $\bm{R} \to \hat{\bm{R}}$ correlation matrices. In addition, we also provide results on previous Basic$^{50}$ and the CRCL without scene labels for comparison, as shown in Fig.~\ref{correlation}. The first column is from the regular event, which shows better orthogonality of the dependent factors compared to the anomalous instances in the second column. The cosine correlations for private and shared features on the same representational dimension (i.e., diagonal elements) are close to 1 while the similarity of the different causal factors (corresponding to non-diagonal elements) is close to 0. The Frobenius Norm (F-Norm) in Table~\ref{t4} also validates this observation, with the F-Norm values showing significant changes in the face of unseen anomalous samples. Therefore, the F-Norm gap between normal and abnormal events suggests that it is feasible to utilize causal representational consistency to perform VAD. In addition, since the ShanghaiTech \cite{shanghai} dataset is collected from various scenarios, neither Basic$^{50}$ nor the CRCL without SdL can overcome the arbitrary scene bias. By viewing columns 1-2 and 3, their F-Norms for regular events are both larger than the full CRCL$^{50}$ model. As analyzed in Sec.~\ref{sec42}, the CRCL that follows the conventional setup without applying scene labels for training is limited by the entanglement representation, so that its performance on the multi-scene dataset is even lower than that of the prior Basic$^{50}$, with an F-Normal gap of 3.5, which is lower than both Basic$^{50}$ and CRCL$^{50}$ with scene-debiasing learning.

\begin{table}[]
  \centering
  \caption{F-Norm values of selected samples from ShanghaiTech.}
  \begin{tabular}{@{}cccc@{}}
  \toprule
  \textbf{Method} & \textbf{Basic$^{50}$} \cite{liu2023learning} & \textbf{CIRL$^{50}$ w/o Scene Labels} & \textit{\textbf{CRCL$^{50}$}} \\ \midrule
  Normal $\downarrow$          & 18.3             & 21.7             & \textbf{15.8}          \\
  Abnormal $\uparrow$        & 22.3             & \textbf{25.2}             & 23.6          \\
  F-Norm GAP $\uparrow$            & 4.0  & 3.5              & \textbf{7.8}           \\ \bottomrule
  \end{tabular}
  \label{t4}
\end{table}

\begin{figure}[t]
  \centering
  \includegraphics[width=\linewidth]{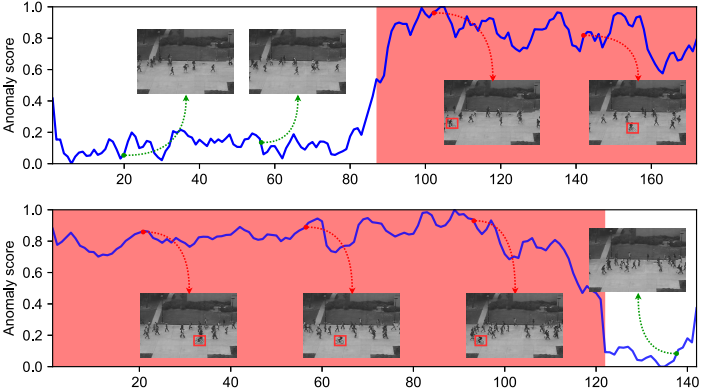}
  \caption{Anomal score curves of selected videos from UCSD Ped2 \cite{ped2} dataset.}
  \label{curve}
\end{figure}

\subsubsection{Anomaly Score Curves} Fig.~\ref{curve} shows the score curves of CRCL$^{50}$ on two sample videos selected from the UCSD Ped2 \cite{ped2} dataset, where the red windows indicate the temporal intervals in which the anomalous events occur. In the VAD community,  the anomaly score curve is a commonly used qualitative evaluation metric for visualizing the model's ability to distinguish anomalies and its response speed. Except for the jitter at the beginning of the video, the anomaly scores calculated by CRCL$^{50}$ consistently remain at a low level during regular time intervals. When the anomaly events occur, the curve rises rapidly and maintains near the high value of about 0.8 until the anomaly ends or leaves the camera's field of view. The results in Fig.~\ref{curve} demonstrate that our proposed CRCL$^{50}$ can quickly respond to video anomalies and accurately localize the time intervals by score curves.

\subsection{Expanded Discussion}

\subsubsection{Validation on Hybrid Dataset} To further validate the effectiveness of the proposed CRCL on surveillance videos with significant scene variations, we hybridize the training sets of UCSD Ped2 \cite{ped2} and CUHK Avenue \cite{avenue} and provided scene labels for the mixed dataset. Existing DeepReL-based methods such as STM-AE \cite{liu2022learning} yield a frame-level AUC of only 68.3\% on the hybrid test set, which is lower than the individual results on either UCSD Ped2 \cite{ped2} or CUHK Avenue \cite{avenue}, suggesting that scene bias cannot be ignored for real-world settings. With the help of scene-debiasing learning, the CRCL$^{18}$ and CRCL$^{50}$ models achieve performance gains of 4.3\% and 2.7\% compared to the previous Basic$^{18}$ and Basic$^{50}$, reaching 77.3\% and 79.1\%, respectively, which significantly outperform the deep learning models. Our proposed SdL utilizes causal inference to explicitly perceive scene biases and remove them with the TDE process, providing a solution for background-agnostic VAD model design.

\begin{table}[]
    \centering
    \caption{Frame-level AUC results of case study on CUHK Avenue \cite{avenue}.}
    \label{tab:cs}
    \resizebox{\linewidth}{!}{
    \begin{tabular}{@{}lcccccc@{}}
    \toprule
    \multicolumn{1}{c}{Proportion} & 1    & 0.9        & 0.8        & 0.7        & 0.6        & 0.5        \\ \midrule
    STM-AE \cite{liu2022learning}  & 89.8 & 88.2\tiny{$_{-1.6}$}     & 86.6\tiny{$_{-3.2}$}     & 84.6\tiny{$_{-5.2}$}     & 83.1\tiny{$_{-6.7}$}     & 79.2\tiny{$_{-10.6}$}    \\
    Basic$^{18}$ \cite{liu2023learning} & 90.5 & 89.6\tiny{$_{-0.9}$} & 89.2\tiny{$_{-1.3}$}     & 87.9\tiny{$_{-2.6}$}     & 87.2\tiny{$_{-3.3}$}     & 85.3\tiny{$_{-5.2}$}     \\
    Basic$^{50}$ \cite{liu2023learning} & 92.5 & 91.3\tiny{$_{-1.2}$} & 90.8\tiny{$_{-1.7}$}     & 89.6\tiny{$_{-2.9}$}     & 88.3\tiny{$_{-4.2}$}     & 87.7\tiny{$_{-4.8}$}     \\
    CRCL$^{18}$  & 91.2 & 90.9\tiny{$_{-0.3}$}   & 90.0\tiny{$_{-1.2}$}     & 89.1\tiny{$_{-2.1}$}     & 88.9\tiny{$_{-2.3}$}     & 87.2\tiny{$_{-4.0}$}     \\
    CRCL$^{50}$ & 92.9 & 92.3\tiny{$_{-0.6}$}   & 92.1\tiny{$_{-0.8}$}     & 91.4\tiny{$_{-1.5}$}     & 91.0\tiny{$_{-1.9}$}     & 90.1\tiny{$_{-2.8}$}     \\  
    \bottomrule
    \end{tabular}
    }
\end{table}

\subsubsection{Case Study of Limited Training Data Available} Current VAD methods that employ DeepReL techniques aim to train models on extensive sets of normal videos to create normality models. These models learn general patterns of regular occurrences and distinguish anomalies by quantifying deviations between test samples and established statistical dependencies. As mentioned in Sec.~\ref{sec1}, such approaches demand a diverse collection of normal samples for effective training. The idea is for the well-trained model to effectively encapsulate a wide array of regular events. However, the real-world scenarios exhibit intricate and diverse spatial-temporal patterns, posing a challenge for training sets to encompass the entire gamut of possible normal distributions. Furthermore, negative samples within the test set may introduce label-independent domain shifts like variations in pedestrian walking speed or clothing color. These shifts can lead to a degradation in performance for conventional methods. To assess the capacity of our proposed CRCL for VAD with limited regular events, we conducted a case study on CUHK Avenue \cite{ped2}. The performance of our CRCL, Basic methods \cite{liu2023learning} and STM-AE \cite{liu2023learning} are examined under conditions where only a portion of training samples are available, as shown in Table~\ref{tab:cs}. Remarkably, our CRCL remains stable performance, even with the absence of a small fraction ($\leq 20\%$) of the training data, with the AUC dropping by a mere 0.6\%. This implies that our approach effectively deduces the underlying normal patterns in test samples and adeptly detects novel anomalies. Even under more challenging setting, where half of the training set is missing, the proposed CRCL$^{18}$ and CRCL$^{50}$ networks exhibit a 2.3\% and 1.9\% frame-level AUC reduction, respectively, lower than the 4.2\% for Basic$^{50}$ and 6.6\% observed in STM-AE \cite{liu2022learning}. The slight degradation suggests that our CRCL is more resilient to label-independent domain shifts that occur in diverse regular events and can maintain stable performance when only limited regular events are available for training.

\section{Conclusion}~\label{sec6}

In this paper, we rethink the unsupervised video anomaly detection task from the causality learning to address the long-standing challenge in deep representation learning that the statistical dependency established cannot cope with label-independent data biases in regular events and detect diverse minor anomalies in the real world. The proposed causal representation consistency learning exploits causal principles to explore potential causal relationships of video normality and detect anomalies with representation consistency, explicitly points out the negative impact of scene bias, and proposes scene-debiasing learning to capture normality-endogenous features, which assists causality-inspired normality learning to mine causal factors. Extensive comparative experiments on three public benchmarks validate the effectiveness of our CRCL and its superiority over various previous work, especially when dealing with multi-scene datasets. The ablation study and extended discussion further show that causal normality learning can resist label-independent data biases such as scene bias and maintain satisfactory performance in practical settings such as multi-scene and limited training samples, providing a deployable solution for VAD towards real-world applications. Video-based anomaly detection systems offer distinct advantages through their broad perceptual coverage and contactless acquisition of spatial-temporal information in real-time. The proposed CRCL framework significantly advances these capabilities by demonstrating superior detection performance and enhanced robustness in challenging real-world scenarios characterized by diverse scenes and limited training data availability. Therefore, it is particularly well-suited for deployment in mission-critical applications such as intelligent transportation systems, automated manufacturing facilities, and smart city infrastructure, where reliable anomaly detection directly contributes to operational safety and efficiency. In the future, we will explore the potential causal mechanisms of multi-view, multi-modal VAD and develop a generalized anomaly detection scheme that can cope with arbitrary working environments and data modality.

\small
\bibliographystyle{IEEEtran}
\bibliography{refs_up.bib}

\end{document}